\documentclass[10pt, a4paper,  notitlepage]{extarticle}

\usepackage[left=2.5cm,right=1cm, top=2.5cm,bottom=2.5cm,bindingoffset=0cm]{geometry}
\usepackage{graphicx}
\usepackage{mathtools}
\usepackage[affil-it]{authblk}
\usepackage{url}
\usepackage{amsmath}
\usepackage{amssymb}
\usepackage{multicol}
\begin{document}
\twocolumn
\title{Sales Time Series Analytics Using Deep Q-Learning}
\author{Bohdan M. Pavlyshenko \\  \small{Ivan Franko National University of Lviv, Ukraine \\ b.pavlyshenko@gmail.com,  www.linkedin.com/in/bpavlyshenko/}}
\maketitle

\subsection*{Abstract}
The article describes the use of deep Q-learning models  in the problems of sales time series 
analytics. In contrast to supervised machine learning which is a kind of passive learning using historical data, 
Q-learning is a kind of active learning with goal to maximize a reward by optimal sequence of actions. 
Model free Q-learning approach for optimal pricing strategies and supply-demand problems was considered in the work. The main idea of the study is to  show that using deep Q-learning approach in time series analytics, the sequence of actions can be optimized by maximizing the reward function when the environment for learning agent interaction can be modeled using the parametric model and in the case of using the model which is based on the historical data. 
In the pricing optimizing case study environment was modeled using sales 
dependence on extras price and randomly simulated demand.
  In the pricing optimizing case study, the environment was modeled using sales dependence on 
  extra price  and randomly simulated demand.  
 In the supply-demand case study, it was proposed to use historical demand time series 
 for environment modeling, agent states were represented by promo actions, previous  demand values and weekly seasonality features. 
 Obtained results show that using deep Q-learning, we can optimize the decision making process for price optimization and supply-demand problems.  Environment modeling using parametric models and historical data can be used for the cold start of learning agent. On the next steps, after the cold start, the trained agent can be used in real business environment.   
 \\
 Key words: Sales, Time Series, deep Q-learning, Reinforcement Learning, Machine Learning.

\subsection*{Introduction}
Sales time series analytics is an important part of modern business intelligence. 
We can mention classical and popular  time series models - Holt-Winters, ARIMA, SARIMA, SARIMAX, GARCH, etc. Different time series models and  approaches can be found in 
~\cite{box2015time, doganis2006time,hyndman2018forecasting,tsay2005analysis}.
In ~\cite{pavlyshenko2019machine} we 
 studied the use of machine-learning models for sales predictive analytics. We considered the main approaches and case studies of using machine learning for sales forecasting, effect of machine-learning generalization. In this paper, we also considered a stacking approach for building regression ensemble of single models. 
In~\cite{pavlyshenko2016linear}, we studied linear models, machine learning and probabilistic models for time series modeling. 
For probabilistic modeling, we considered the use of copulas and Bayesian inference approaches. 
In~\cite{pavlyshenko2016machine}, we studied the logistic regression in the problem of  detecting manufacturing failures. 
For the logistic regression, we considered a generalized linear model, machine learning  and Bayesian models. 
In~\cite{pavlyshenko2018using}, we studied stacking approaches for time series forecasting and logistic regression with highly imbalanced data. 
The use of regression approaches for sales forecasting can often give us better results compared to time series methods. One of the main assumptions of regression methods is that the patterns in the historical data will be repeated in future. 
Supervised machine learning can be considered as a kind of passive learning using historical observations. 
Time series forecasting using Machine Learning gives us insights and allows us to make right business decisions.
Reinforcement learning allows us to find sequences of optimized actions directly without historical data. 
In this approach, we have an environment and a learning agent which interacts with environment.
As a result of each interaction, learning agent receives reward. The goal of Reinforcement 
Learning  is to find such sequence of actions which will maximize an average cumulative  reward on the episodes of agent-environment interactions. There are policy based and policy free approaches. Policy can be described by parametrized distribution function for states and actions. 
The parameters of these distributions can be found using policy gradient approach where on each iteration, we calculate the gradient of objective function. 
Policy free approach can be Q-learning which is based on Bellman equation ~\cite{sutton1998introduction, mnih2015human, mnih2013playing} .
On each iteration, we upgrade the Q-table where rows represent states and columns represent actions. 
In the case of continuous action, space Q-table can be approximated by Neural Network using DQN approach ~\cite{mnih2015human, mnih2013playing}.
\subsection*{Related Work}
The main principles of reinforcement learning can be found at
~ \cite{sutton1998introduction}. In ~\cite{mnih2015human, mnih2013playing} deep Q-learning approaches were studied. 
In ~ \cite{rana2014real}, real-time dynamic pricing in a non stationary
environment has been considered. In the article, the problem of establishing a pricing policy that maximizes the revenue for selling a given inventory by a fixed deadline has been considered.
In ~ \cite{maestre2018reinforcement} reinforcement learning for fair dynamic pricing has been considered.
In ~ \cite{vengerov2007gradient} , 
 Reinforcement Learning algorithm that can tune parameters of a seller’s dynamic pricing policy in a gradient direction  even when the seller's environment is not fully observable was considered.
Different approaches for dynamic pricing is considred in ~ \cite{den2015dynamic}. In the paper ~\cite{kim2005adaptive} adaptive inventory-control models for a supply chain consisting of one supplier and multiple retailers were proposed. The paper ~\cite{ raju2003reinforcement} describes 
 the use of reinforcement learning techniques in the problem of determining dynamic prices in an electronic retail market. The papers ~\cite{huang2018financial, jiang2017deep} consider the use of reinforcement learning for financial  analytics.  
The paper ~\cite{liu2005neural}  formulates an autonomous data-driven approach to identify a parsimonious structure for the NN so as to reduce the prediction error and enhance the modeling accuracy. The Reinforcement Learning based Dimension and Delay Estimator (RLDDE) was proposed. The paper ~\cite{jiang2017deep} presents a model-free Reinforcement Learning framework to provide a deep machine learning solution to the portfolio management problem. In ~\cite{dulac2015deep}, 
Deep Reinforcement Learning in Large Discrete Action Spaces was considered.

\subsection*{Deep Q-Learning Approaches}
The goal of Q-learning is to maximize cumulative future reward ~\cite{sutton1998introduction,mnih2015human, mnih2013playing}. 
To training the Q-learning network, the gradient descent algorithm is often used. 
To eliminate influence between sequence data and non-stationary  distribution,
the replay mechanism ~\cite{lin1993reinforcement} can be used.
This approach consists in random sampling of previous data which represent 
states and actions.  It makes it possible to average distributions of data which describe previous agent's behaviour. The goal for the agent lies in choosing a sequence action strategy which maximizes future rewards ~\cite{mnih2013playing}. 
An optimal action-value function can be considered as:
\begin{equation}
Q^*(s,a)=\mathbb{E}_{s' \sim \varepsilon}\left[ r+ \gamma \underset{a'}{\max}Q(s',a') 
\mid s,a \right] 
\end{equation} 
where $r$ is a reward, $s$ is a state, $a$ is an action, $s', a'$ are possible states and actions on the next time step.
To estimate the function $  Q^*(s,a) $ approximately, Bellman equation can be used in the iteration process:
 \begin{equation}
Q_{i+1}(s,a)=\mathbb{E}\left[ r+ \gamma \underset{a'}{\max}Q_{i}(s',a') 
\mid s,a \right] 
\end{equation} 
The main problem of such an approach is that there is no generalization of revealed patterns in the agent-environment interaction due to the fact that 
 $  Q(s,a) $ is estimated on each separate step. 
 To improve generalization, one can use an approximation function for 
 $  Q^*(s,a) $.  Neural network can be used for this purpose. 
 Parameters of such deep Q-network can be found with gradient methods minimizing 
 the loss function
 \begin{align}
 &L_i(\theta_i)=\mathbb{E}_{s,a \sim \rho}\left[ (y_i-Q(s,a;\theta_i))^2 \right] \\
 &y_i=\mathbb{E}_{s' \sim \varepsilon}\left[ r+ \gamma \underset{a'}{\max}Q(s',a',\theta_{i-1}) 
\mid s,a \right] 
\end{align} 
where $  \rho $ - is the behaviour distribution, $\theta$ is the weights of Q-network.

Experience replay  approach ~\cite{lin1993reinforcement}  is effectively used in deep Q-network ~\cite{mnih2013playing,mnih2015human}.
In this approach, agent actions and states are stored in the replay memory at each time step as tuples 
$e_t=(s_t,a_t,r_t,s_{t+1})$. Tuples $e_t$ are stored in the data set $D_t=\lbrace e_1,...e_t \rbrace$.
On each step for Q-learning updates, the algorithm gets samples $e_t$ from the replay memory by uniform random sampling $e_t \sim U(D)$ ~\cite{mnih2013playing}. 
  On the next experience replay step, Q-learning updates  
the weights $\theta$. On the next step, the  agent selects an optimal action, using $ \varepsilon $-greedy policy ~\cite{mnih2013playing}.   
Data mini batches are formed on each iteration for updating the weights for 
Q-network.
Mini batches are chosen randomly. 
Such an approach provides generalization of approximation given  data for agent-environment interaction. 
Due to the experience replay approach, behavioural distribution is averaged 
on many previous states and actions of learning agent that provides the convergence of iteration process. 
One of widely used approaches for Q-network consists in treating agent states as input parameters when outputs are $Q$-values for each separate agent action ~\cite{mnih2013playing}. 
In such an approach,   $Q$-value for each action is calculated during one Q-network  step forward. 

In this work, we consider two cases of using Q-learning in sales time series analytics. 
One case is an optimal pricing strategy, the second case is supply and demand problems which appear in retail domain areas. In our numerical experiments, we used the algorithms which were based on model architecture described in ~\cite{mnih2015human, mnih2013playing} and the approaches for Q-learning agent implementations from ~\cite{githubrep1,githubrep2,githubrep3}. For environment modeling,
we used the parametric models in the case of price optimization and historical time series of demand in the case of supply-demand problem.  
For the analysis of supply-demand problem, we used the store sales historical data from  'Rossmann Store Sales' Kaggle competition ~\cite{rossmanstorekaggle}. These data describe sales in Rossmann stores.  The calculations were conducted in the Python environment using the main packages \textit{pandas, sklearn, numpy, keras, matplotlib, seaborn}.  To conduct the analysis, \textit{Jupyter Notebook} was used.

\subsection*{Optimal  Pricing Strategy using Q-Learning}
\begin{figure}
\centerline{\includegraphics[width=0.3\textwidth]{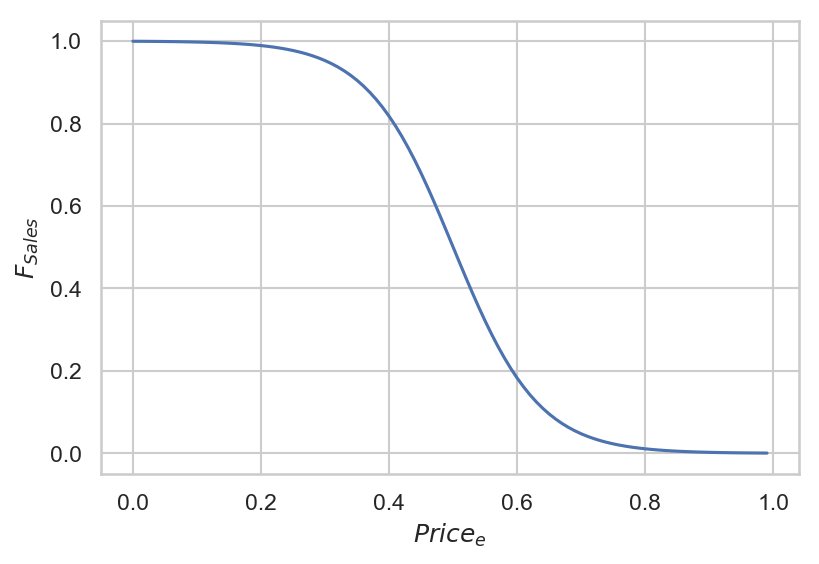}}
\caption{Sales vs extra prices}
\label{fig1}
\end{figure}

Let us consider a simple case for the pricing strategy.  
The question is what strategy can be applied to maximize profit for some time period. 
Generally, the strategy can consist of many factors and means. In the simplest model, it consists of discrete extra price values which are the fraction of cost price. Dependence between sales and extra price can be considered as
 \begin{equation}
 F_{Sales}=\frac{a}{(1+b \cdot exp(c \cdot( Price_m\cdot(1+Price_{e})-d)))},
 \label{opt_price_f1}
 \end{equation}
 where $Price_m$ is an marginal price, $Price_{e}$ is an  extra price, $a,b,c,d$ are the  parameters for 
 function $F_{Sales}$. 
 The function $F_{Sales}$ ~(\ref{opt_price_f1}) describes relative decreasing of sales with an increasing extra price. Extra price is considered in relative parts of marginal price.   
 The reward function for Q-learning can be considered as 
 \begin{equation}
 Reward=Demand \cdot  F_{Sales} \cdot Price_e
  \label {opt_price_f2}
 \end{equation}
We created a simple model with normalized variables. The Figure~\ref{fig1}  shows sales vs extra prices ($Price_m$=1, a=1, b=1, c=15, d=1.5).
We can observe that for high extra price we get small sales. The real parameters for the logistic curve can be found by the gradient method using historical data. 
\begin{figure}
\centerline{\includegraphics[width=0.3\textwidth]{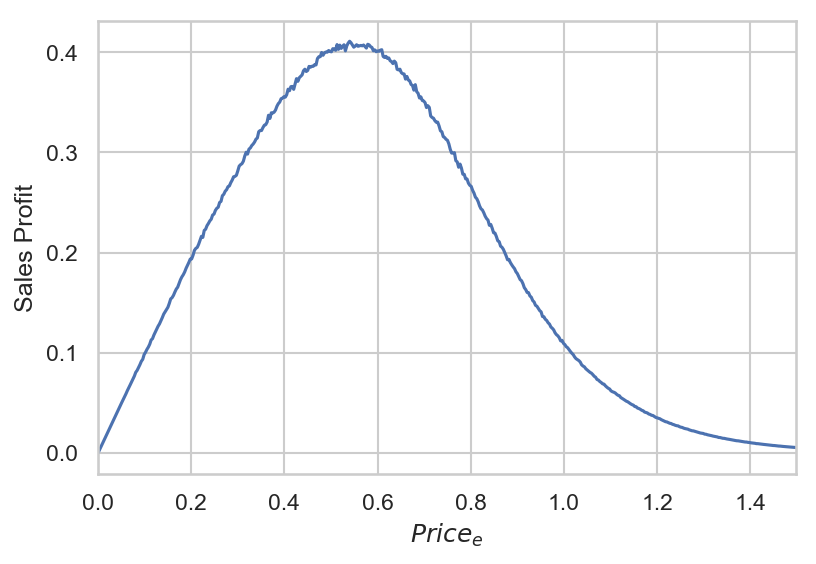}}
\caption{Profit versus extra prices}
\label{fig2}
\end{figure}
\begin{figure}
\centerline{\includegraphics[width=0.5\textwidth]{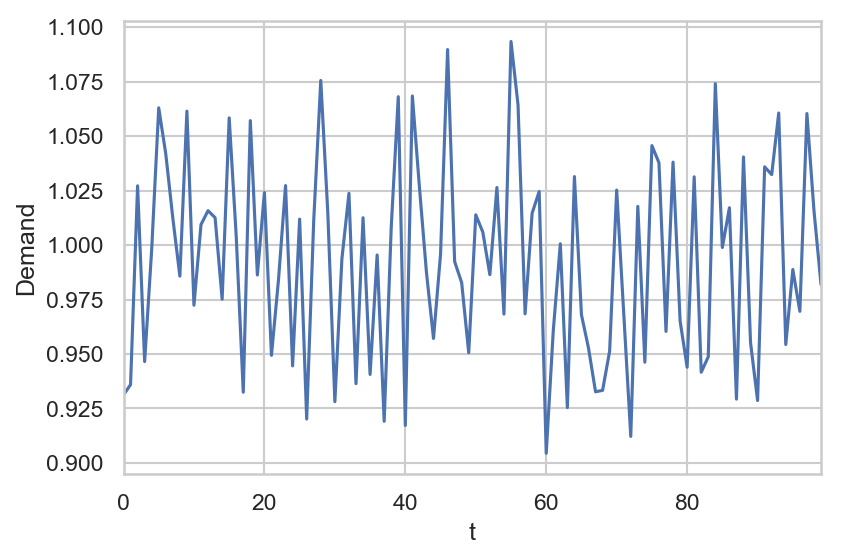}}
\caption{Simulated demand time series.}
\label{fig3}
\end{figure}
 The Figure  ~\ref{fig2}  shows profit versus extra prices.   One can see that there is no profit with low and high extra price.
 
The challenge is to find the optimal extra price. We can tabulate the objective function using simulated demand time series and find  the optimal value for extra price.
But in real cases, profit versus extra prices can have a complicated functional dependence, including the dependence on many qualitative factors which are included into complex multilevel 
pricing strategy. To find which pricing strategy is optimal, we can use a Q-learning approach. 
This simple model can be used for the cold start of learning agent before the interaction with real business environment.

Let us consider the parameters for numerical modeling.
For the Q-values approximation we used 
2-layers feed forward neural network with 32 neurons in each 
layer. Output neural network dimension is equal to the number of possible actions, which was chosen equal to  8.  For the actions, we took the list of the  following  values for normalized extra price [0, 0.15, 0.25, 0.5, 0.75, 0.85, 1, 1.5].
The extra price is considered in relative parts of the marginal price.  
As time steps we consider days.  
The number of time steps in the each episode equal to 7, the batch size was 32,
the number of learning iterations was 50, the optimizer is Adam, the learning rate for neural network was 0.001, the epsilon decay was 0.97.  For approximation of the function (~\ref{opt_price_f1}), we used the following parameters: $Price_m$=1, a=1, b=1, c=7, d=1.7

\begin{figure}
\centerline{\includegraphics[width=0.3\textwidth]{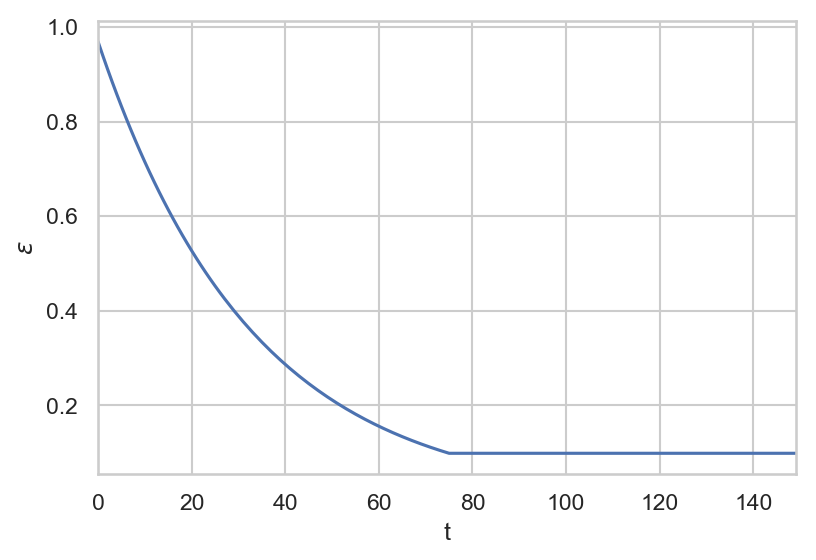}}
\caption{Epsilon vs time dependence}
\label{fig4}
\end{figure}

\begin{figure}
\centerline{\includegraphics[width=0.5\textwidth]{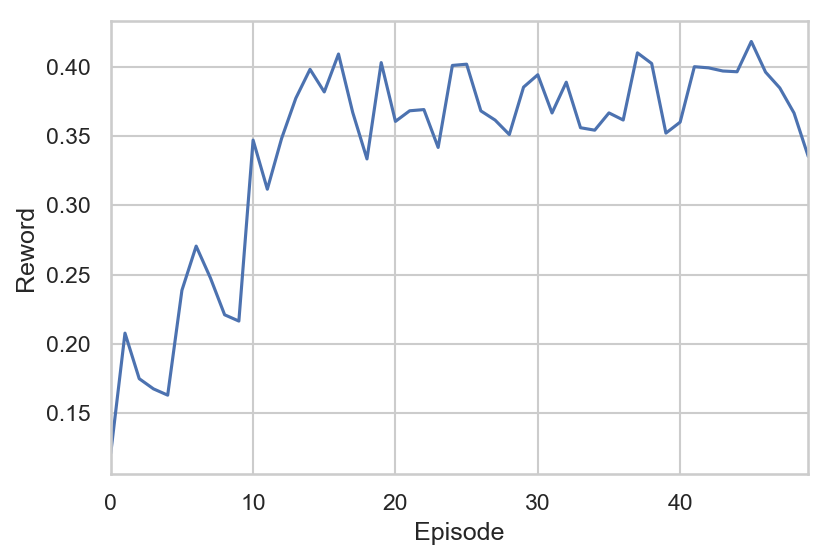}}
\caption{Mean reward over episodes}
\label{fig5}
\end{figure}

\begin{figure}
\centerline{\includegraphics[width=0.5\textwidth]{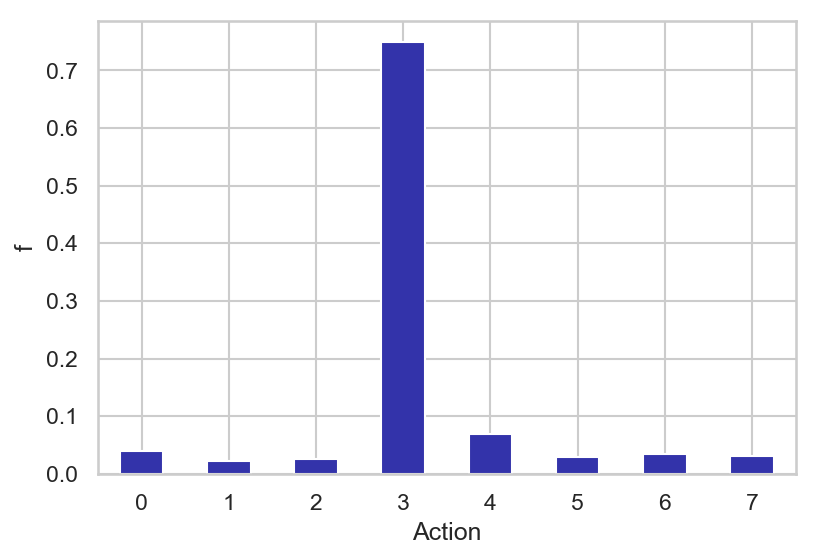}}
\caption{Frequencies of actions}
\label{fig6}
\end{figure}

\begin{figure}
\centerline{\includegraphics[width=0.5\textwidth]{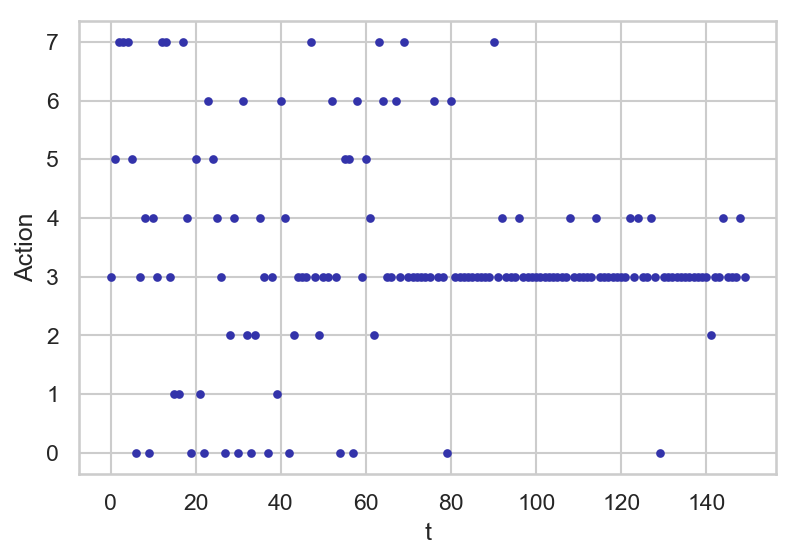}}
\caption{ Actions vs time}
\label{fig7}
\end{figure}

In this case study, we used DQN approach with epsilon-greedy exploration-exploitation trade off. 
Epsilon describes the probability of random action. On each iteration, epsilon decreases. 
The Figure ~\ref{fig4} shows epsilon vs time dependence. For the numerical experiment,  we simulated demand with random uniform distribution of relative units. As a  result of numerical experiment we received trained DQN model which can output optimal actions that maximize future cumulative reward. The Figure ~\ref{fig3}  shows simulated demand time series. The Figure ~\ref{fig5}  shows mean reward over the episodes.
One can observe that the reward  goes up with iterations. It means that the learning agent improves the way it interacts with the environment. 
The Figure ~\ref{fig6} shows the frequencies of actions and we can see one dominated action which corresponds to the optimal pricing strategy for this simple model.
The Figure ~\ref{fig7} shows the actions with time and we can see a big dispersion of actions at the beginning  of interaction process due to the domination of exploration type of interactions. 
We can observe that on the next steps after exploration time period, one action dominates which can be considered as a found optimal pricing strategy.

\subsection*{Supply and Demand Case Study using Q-Learning}
\begin{figure}
\centerline{\includegraphics[width=0.5\textwidth]{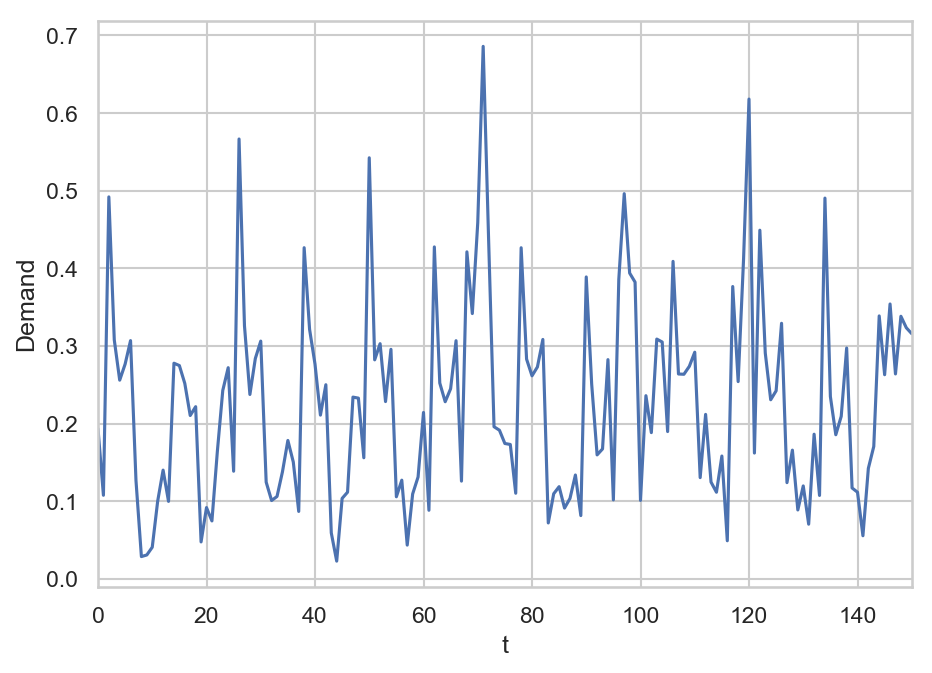}}
\caption{Simulated demand time series}
\label{fig8}
\end{figure}

\begin{figure}
\centerline{\includegraphics[width=0.5\textwidth]{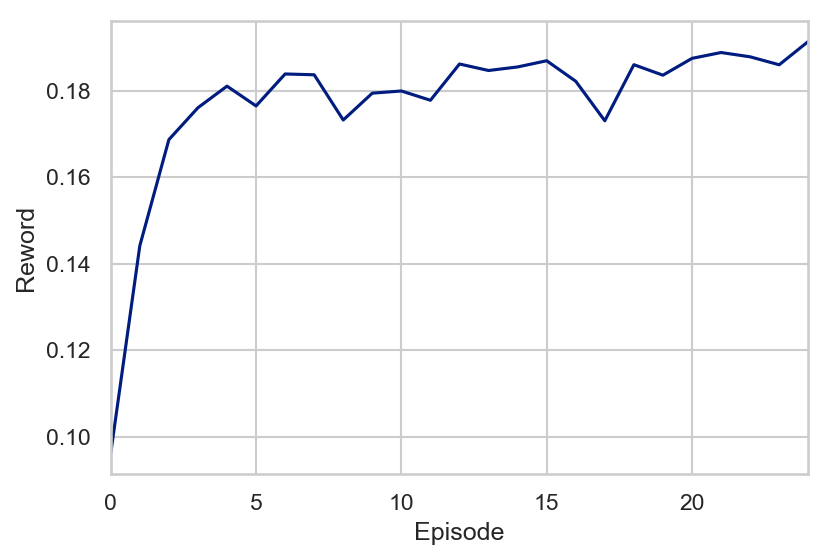}}
\caption{Mean reward over episodes}
\label{fig9}
\end{figure}

Let us consider a case study of using Q-learning for supply and demand problems. 
We can also use historical data for the initial start of Q-learning algorithm along with parametric modeling of the environment. Such an approach makes it possible to conduct the cold start for Q-learning agents. In this case study, we took normalized demand time series with seasonality and promo action factor.
The challenge is to find optimal discrete actions in the supply-demand problem. Products can be supplied by batches with discrete amount. In the model, we took into account the  expenses for product processing which are related to  logistic, storing and other expenses.  
The reward on each step can be considered as 
\begin{equation}
 Reward=SalesProfit -  ProcessCost
\end{equation}
In this case study, we have more features for state representation.
They are  the product demand the day before, expected promo action, day of week. 
Let us consider the parameters for modeling supply-demand problem.
For modeling environment, we used the historical time series for demand. 
For each episode, the demand for 150 days was taken. 
This time series was simulated given the sales time series from   'Rossmann Store Sales' Kaggle competition ~\cite{rossmanstorekaggle}.
 To eliminate overfitting, random lag for the first point of demand time series was applied. The lag was calculated by uniform random distribution for integer values ranging from 0 to 25.
 For each episode, we used  the  time period of 150 days, the number of actions was 7. We used the feed forward neural network with 2 layers with 64 neurons in each layer. The batch size was 32, learning rate was 0.001, epsilon decay was 0.995, gamma coefficient  for Bellman equation was 0.3. To calculate the reward, we used the following parameters: the value of price profit was 1, pack unit was 0.05,  price support was 0.5. 
For state features we used a promo binary feature, 
previous day sales, weekly seasonality features which were 
7 binary features for each week day which were obtained by one hot encoding 
of day of  week categorical feature. 
We also set up generating stop episode flag when the reward became lower than    specified reward. 
Applying this rule accelerates the process of Q-learning, since in the cases 
of low current reward values learning process were stopped and new learning 
iteration was started.   
These actions were considered as discrete values for supply. For numerical modeling, 7 actions were chosen  with the following values [ 0,  2,  4,  6,  8, 10, 12], which are a discrete number of packs. The amount of simulated product in the pack was 0.025 of relative units. 
 The Figure ~\ref{fig8} shows simulated demand time series for arbitrary chosen episode.
 For numerical experiment, demand is simulated in some relative units. 
The Figure ~\ref{fig9} shows calculated mean reward over episodes.

\begin{figure}
\centerline{\includegraphics[width=0.5\textwidth]{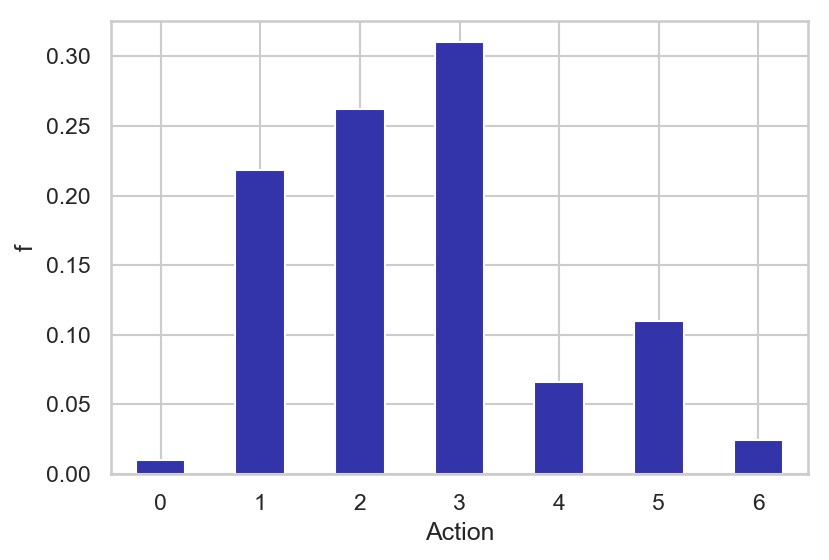}}
\caption{Frequencies of actions for supply-demand problem}
\label{fig10}
\end{figure}

\begin{figure}
\centerline{\includegraphics[width=0.5\textwidth]{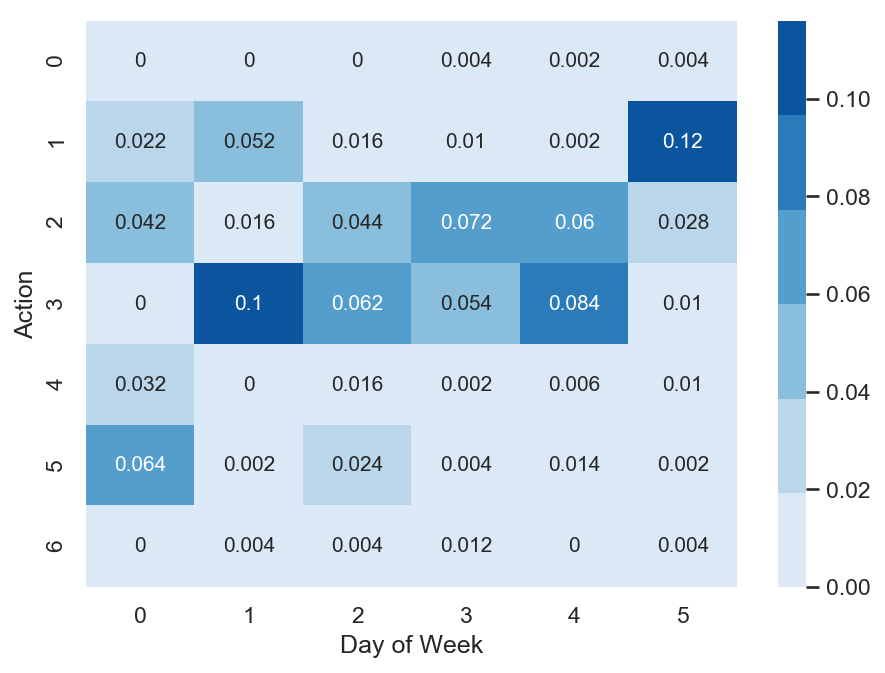}}
\caption{ Heatmap for action frequencies vs weekday}
\label{fig11}
\end{figure}

We can observe that the learning agent optimizes action sequences over the episodes. In Figure ~\ref{fig10}, we can see that several actions dominate in comparison with the previous case study for price strategy optimization where only one 
action dominated.
Here we have more features for state representation, so the learning agent chose different optimal actions for different states. 
The Figure ~\ref{fig11} shows  the heatmap for action frequencies against weekday. 
\begin{figure}
\centerline{\includegraphics[width=0.5\textwidth]{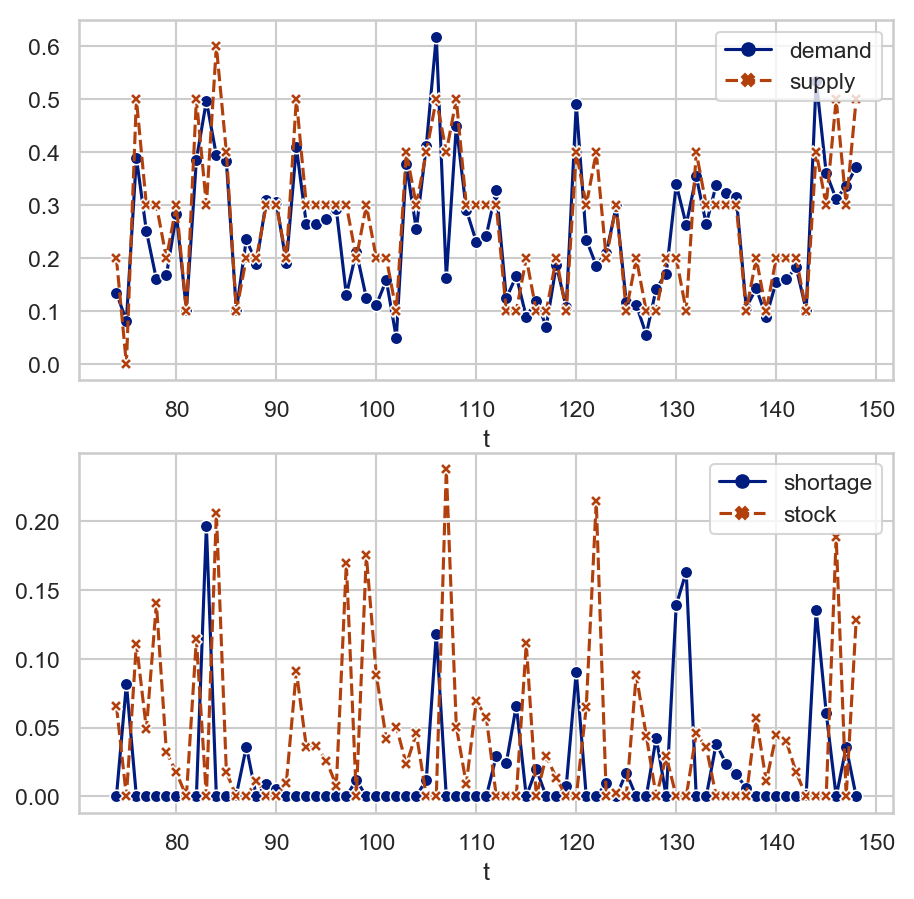}}
\caption{Time series of demand, supply, stock and shortage }
\label{fig12}
\end{figure}
We can see that for different days we have different dominated actions which also depend on promo action which  can take place in different days. The Figure ~\ref{fig12} shows  the time series of demand, supply, stock and shortage.
Dynamics of supply, shortage and stock depend on reward function which can be formed by by profit, expenses on logistic,  product processing and lost profit.  

\subsection*{Conclusions}
The article describes the use of deep learning models for Q-learning in the problems of sales time series 
analytics. In contrast to supervised machine learning which is a kind of passive learning using historical data, 
Q-learning is a kind of active learning with goal to maximize the reward by optimal sequence of actions. 
Model free Q-learning approach for optimal pricing strategies and supply-demand problems was considered in the work. It was shown that using deep Q-learning 
approach, the sequence of actions can be optimized by maximizing the reward function. 
  In the pricing optimizing case study, the environment was modeled using sales dependence on 
  extra price  and randomly simulated demand.  
 In the supply-demand case study, it was proposed to use historical demand time series 
 for environment modeling, agent states were represented by promo actions, previous  demand values and weekly seasonality features. 
 Obtained results show that using deep Q-learning, we can optimize the decision making process for price optimization and supply-demand problems.  Environment modeling using parametric models and historical data can be used for the cold start of learning agent. On the next steps, after the cold start, the trained agent can be used in real business environment.   
 So, using Q-learning we can build a decision-making algorithm, which proposes qualitative decisions. This algorithm can start 
with the data simulated by parametric expert model or with the model based on historical data and then it can work with real business environment and do adaptation to the  changes in business processes. 
In more complicated cases, we can take into account the price for spoilt products in case we deal with perishable products and other expenses.

\bibliographystyle{ieeetr}
\bibliography{ref}
\end{document}